%% file: main.tex
\documentclass{esannV2}
\usepackage[dvips]{graphicx}
\usepackage[latin1]{inputenc}
\usepackage{amssymb,amsmath,array}
\usepackage[ruled,vlined,linesnumbered]{algorithm2e}
\usepackage{booktabs}
\usepackage[table,xcdraw]{xcolor}
\usepackage{hyperref}

\input{commands}

\begin{document}
\title{Improving Fast Minimum-Norm Attacks with Hyperparameter Optimization} 

\author{Giuseppe Floris$^{1*}$ , Raffaele Mura$^{1*}$, Luca Scionis$^{1*}$, \\ Giorgio Piras$^{1, 2}$, Maura Pintor$^1$, Ambra Demontis$^1$ and Battista Biggio$^1$
%
\vspace{.3cm}\\
%
1- University of Cagliari - Department of Electrical and Electronic Engineering 
%
\vspace{.1cm}\\
2- Sapienza University of Rome - Department of Computer Engineering \\
}

\maketitle
\def\thefootnote{*}\footnotetext{These authors contributed equally to this work}

\begin{abstract} 
Evaluating the adversarial robustness of machine-learning models using gradient-based attacks is challenging. In this work, we  show that  hyperparameter optimization can improve fast minimum-norm attacks by automating the selection of the loss function, the optimizer, and the step-size scheduler, along with the corresponding hyperparameters. Our extensive evaluation involving several robust models demonstrates the improved efficacy of fast minimum-norm attacks when hyped up with hyperparameter optimization. We release our open-source code at \url{https://github.com/pralab/HO-FMN}.

\end{abstract}

\section{Introduction} 
Machine learning (ML) models are susceptible to adversarial attacks~\cite{biggio13-ecml, szegedy14-iclr}, i.e., input samples carefully perturbed to mislead the model. 
To evaluate adversarial robustness, many different gradient-based attacks have been proposed, whose performance is significantly affected by the choice of the loss function to optimize, the optimization algorithm, and the step-size scheduler.
From a practical perspective, attacks tend to be run with a ``default'' configuration and set of hyperparameters that are deemed to fit most of the cases. 
Yet, the attack effectiveness is highly case-dependent, implying that the choice of the configuration needs to be carefully tailored to the model rather than a de-facto standard choice. 
In AutoAttack (AA)~\cite{croce20-autoattack}, the authors try to overcome this limitation by proposing an ensemble of parameter-free attacks, each including an internal auto-tuning process for each relevant hyperparameter. With Adaptive AutoAttack (AAA)~\cite{aaa_yao20}, the approach is configured to run the parameter-free AA to look for a fast and good evaluation or alternatively come forward with an extensive search on a pool of attacks.

In this paper, we aim to use a smart and effective search for the best configuration that adapts the attack to the model. 
Hence, we propose a systematic framework for configuring the state-of-the-art, fast minimum-norm (FMN) attacks properly instead of running extensive searches on multiple attacks. 
To this end, we develop our framework by rethinking the choice of the loss function, optimizer, and step-size scheduler as attack hyperparameters and then using a unified hyperparameter optimization procedure.


\section{FMN Attacks with Hyperparameter Optimization}\label{sect:fmn_ho}

We introduce here a modified FMN attack algorithm, referred to as HO-FMN, in which the loss function, the optimizer, and the step-size scheduler, along with their hyperparameters, are all exposed to be optimized. We then provide details on the hyperparameter optimizer considered in this work.

\myparagraph{FMN Attacks.} FMN~\cite{pintor21-neurips} aims to find minimum-norm adversarial perturbations. The objective of the attack is to find, for a model with decision function $f(\cdot)$, the smallest perturbation $\vct \delta$ to add to an input sample $\vct x$ with true label $y$ so that $f(\vct x + \vct \delta) \neq y$. To this end, it takes a PGD-step to optimize the perturbation by minimizing a loss function (i.e., the {$\delta$-step}) within a given perturbation budget $\epsilon$, then it adjusts $\epsilon$ to iteratively reduce the perturbation norm (i.e., the {$\epsilon$-step}).
In this work, we improve the $\delta$-step in the FMN algorithm while leaving the $\epsilon$-step unchanged (as the latter only optimizes a scalar value).
In this regard, we consider different step-size schedulers (other than the \textit{cosine annealing}~\cite{loshchilov17-SGDRwarm} used in the baseline FMN) and more sophisticated optimizers (using different gradient update strategies other than SGD). The attack loss on which we optimize (logit loss \cite{carlini17-sp}) is also put into question as we look for better candidates.

\myparagraph{HO-FMN.}\label{paragraph:ho}
We report in Algorithm~\ref{alg:alg_fmn} a revisited formulation of the FMN attack, in which the role of the loss function $L$, optimizer $u$, and scheduler $h$ are better isolated. 
This novel formulation of the FMN attack enables us to generalize it by allowing a different selection of each component, treating each of them as a different hyperparameter or attack configuration. 
\input{algorithm} 
While the overall algorithm remains conceptually unchanged, we modify the attack loss $L$, the optimizer $u$, and the step-size scheduler $h$ used in the $\delta$-step. These are the elements that change from the original attack implementation and are made optimizable in our work. In practice, given a model, we exploit hyperparameter optimization to find the best combination of loss, optimizer, and scheduler along with their hyperparameter values (e.g., the initial step size, etc.).

\section{Experiments}\label{sect:exp}

We describe below the experimental setup used in our work and then the results. 

\myparagraph{Hyperparameter Tuning.}
As introduced in Sect.~\ref{sect:fmn_ho}, we aim to improve FMN by optimizing the choice of: 
(i) the loss function, selecting between the logit loss (LL) and the cross-entropy loss (CE); 
(ii) the optimizer, selecting between SGD (with and without Nesterov acceleration) and Adam (with and without AmsGrad); and 
(iii) the step-size scheduler, selecting among Cosine Annealing (CALR), Cosine Annealing with Warm Restarts (CAWR), MultiStep (MSLR), and Reduced On Plateau (RLROP).
We define for optimizers and step-size schedulers a search space made by the possible hyperparameter values and sampling options. In particular, for each optimizer, we tune the initial step size, the momentum, and weight decay; for each scheduler, we tune the most important parameters such as the milestones in MSLR, the iteration parameters in CALR and CAWR and the factor in RLROP. 
The search space is then parsed in the context of the FMN hyperparameter optimization. The algorithms responsible for finding the best optimizer/scheduler configuration are the CFO search algorithm~\cite{wu2020AAAI-cfo} and the ASHA scheduler~\cite{li2020-asha}. We leverage the Ray Tune framework\footnote{\url{https://docs.ray.io/en/latest/tune/index.html}} for handling the hyperparameter optimization~\cite{liaw2018tune}.

\myparagraph{Dataset.}
We take a subset of $100$ samples from the CIFAR10 test set for running our hyperparameter optimization, where for every model we analyzed each and every loss/optimizer/scheduler configuration. Upon finding a specific set of best hyperparameters, we use a separate set of $1000$ samples (also taken from the CIFAR10 test set) to run the FMN attack on the models and discuss the results.

\myparagraph{Perturbation Model.} We restrict our analysis here to the $\ell_\infty$-norm attacks, as it is one of the most problematic cases for the baseline FMN algorithm.

\myparagraph{Models.}
We consider 9 state-of-the art robust models from RobustBench~\cite{croce20-robustbench}: 
\textit{M0}, the WideResNet-70-16 in~\cite{wang23-better}; 
\textit{M1}, the WideResNet-28-1 in~\cite{wang23-better}; 
\textit{M2}, the WideResNet-70-16 in~ \cite{gowal21-improving}; 
\textit{M3}, the WideResNet-106-16 in~\cite{rebuffi22-fixing}; 
\textit{M4}, the WideResNet-28-10 in~\cite{gowal21-improving}; 
\textit{M5}, the WideResNet-70-16 in~\cite{pang22-robustness}; 
\textit{M6}, the ResNet-152 in~\cite{sehwag22-proxy}; 
\textit{M7}, the WideResNet-28-10 in~\cite{pang22-robustness}; and
\textit{M8}, the ResNet-18 in~\cite{gowal21-improving}. 

\myparagraph{Performance Metrics.} 
For HO-FMN, we select the configuration that achieves the smallest median $||\vct \delta||_\infty$, following~\cite{pintor21-neurips}. We then use it to evaluate the robust accuracy (RA) of the models. For minimum-norm attacks, RA can be evaluated over the entire range of perturbation sizes by imposing a threshold on the distances found, i.e., we can compute the full robustness-perturbation curve.

\myparagraph{Experimental Results.} We evaluate the candidate configurations by running HO-FMN on the 9 selected models. We show the resulting robust accuracy values at $\epsilon=8/255$ in \autoref{tab:configurations}. We highlight how the LL loss consistently finds better perturbations than CE and that, in the case of the most unfortunate configuration, the models' robustness would have been considerably overestimated. 

\input{table}

From the results in \autoref{tab:configurations} we select the configurations Adam+RLROP+LL and SGD+CALR+LL, and run the evaluation on our test set of 1000 CIFAR10 samples with the discovered best hyperparameters.\footnote{\url{https://pytorch.org/docs/stable/optim.html}} 

We report the hyperparameters found with the first setting, which we name HO-FMN (Adam), listed for each model as  (\texttt{learning\_rate}, \texttt{weight\_decay}, \texttt{factor}, \texttt{amsgrad}), 
M0: (5,534	0,025	0,327, False); 
M1: (8,801	0,043	0,366, False); 
M2: (4,073	0,019	0,286, False); 
M3: (9,616	0,024	0,301, False); 
M4: (7,078	0,019	0,260, False); 
M5: (7,078	0,019	0,260, False); 
M6: (4,194	0,020	0,235, False); 
M7: (9,339	0,023	0,352, True); 
M8: (4,073	0,019	0,286, False).
We leave the other parameters of Adam+RLROP fixed: \texttt{eps}=$10^{-8}$, \texttt{betas}=(0.9, 0.999), \texttt{patience}=5, \texttt{threshold}=$10^{-5}$.

We list the hyperparameters for the second configuration, HO-FMN (SGD), as (\texttt{learning\_rate}, \texttt{weight\_decay}, \texttt{momentum}, \texttt{dampening}),
M0: (4,453	0,917	0,010	0,085); 
M1: (1,523	0,880	0,041	0,037); 
M2: (1,222	0,916	0,010	0,089); 
M3: (3,837	0,924	0,001	0,114); 
M4: (1,013	0,926	0,014	0,122); 
M5: (3,141	0,936	0,010	0,136); 
M6: (1,222	0,943	0,010	0,058); 
M7: (2,562	0,911	0,010	0,071); 
M8: (2,124	0,922	0,010	0,104),
and we keep fixed \texttt{T\_max}=100, \texttt{eta\_min}=0, \texttt{last\_epoch}=-1.

\begin{figure*}[htbp]
\centering

    \includegraphics[width=\textwidth]{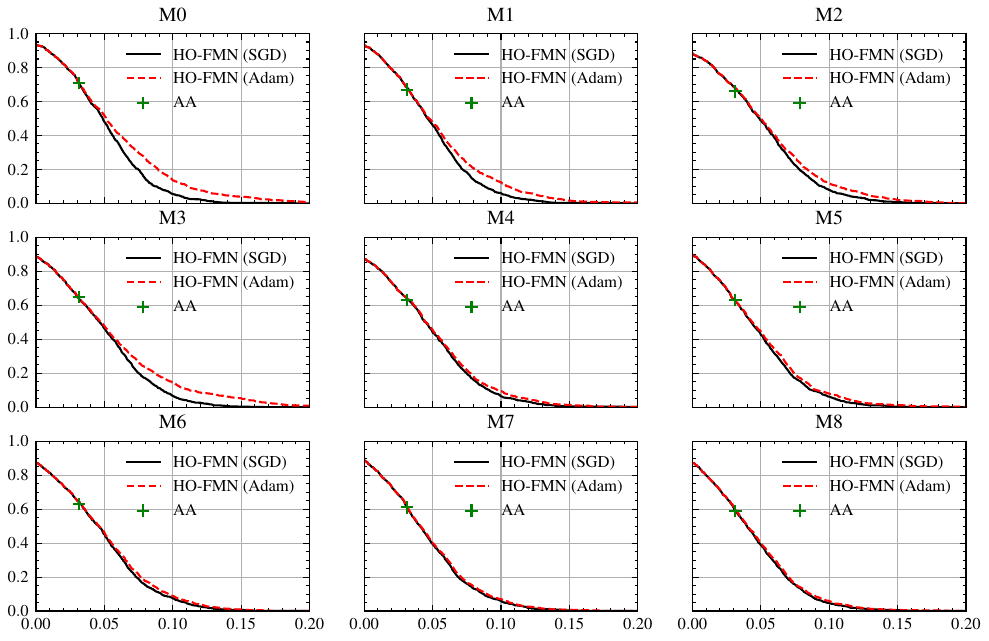}

    \caption{Robust accuracy computed on M0-M8 with HO-FMN in the two versions Adam and SGD, and with AA at the fixed value of $\epsilon$=8/255.}\label{fig:sec_evals}
\end{figure*}

We show the resulting robust accuracy for increasing perturbation sizes in \autoref{fig:sec_evals}, compared with the value reported in RobustBench (computed only for $||\vct \delta || = 8 / 255$). We remark that the values found are comparable with AA. However, our attack is able to compute the full robustness-perturbation curve with one single optimization. Achieving the same result with AA is only possible by running AA multiple times with a range of different perturbation sizes, which would require a significant increase in computational complexity.

\section{Conclusions and Future Work} 
In this work, we investigated the use of hyperparameter optimization to improve the performance of the FMN attack algorithm. Our findings highlight that hyperparameter optimization can improve FMN to reach competitive performance with AutoAttack while providing a more thorough adversarial robustness evaluation (i.e., computing the whole robustness-perturbation curve).

We argue that the same approach can be combined with other attacks, different threat models, or more configurations. 
In future work, we will extend our analysis beyond the $\ell_\infty$-norm FMN attack, considering  $\ell_0$, $\ell_1$, and $\ell_2$ norms. 

We remark that adding more hyperparameters to tune would make the search space bigger, resulting in a longer optimization time. To this end, we will also try to develop sound heuristics to make hyperparameter tuning faster, designing faster exploration phases in the initial steps of the FMN optimization process.




\section*{Acknowledgements}
We would like to express our gratitude to Sarthak Gupta for his preliminary experiments that contributed to the realization of this project.
This work has been carried out while G. Piras was enrolled in the Italian National Doctorate on AI run by the Sapienza University of Rome in collaboration with the University of Cagliari. It has been supported by the PRIN 2017 project RexLearn (grant no. 2017TWNMH2), funded by the Italian Ministry of Education, University and Research; by BMK, BMDW, and the Province of Upper Austria in the frame of the COMET Programme managed by FFG in the COMET Module S3AI; and by Fondazione di Sardegna under the project ``TrustML: Towards Machine Learning that Humans Can Trust'', with CUP: F73C22001320007.


\begin{footnotesize}

\bibliographystyle{abbrv}
\bibliography{biblio}

\end{footnotesize}


\end{document}

%% file: commands.tex
\newcommand{\myparagraph}[1]{\noindent \textbf{#1}}

\newcommand{\vct}[1]{\ensuremath{\boldsymbol{#1}}}

%% file: algorithm.tex
\begin{algorithm}[t]
 \SetKwInOut{Input}{Input}
    \SetKwInOut{Output}{Output}
    \SetKwComment{Comment}{$\triangleright$\ }{}
    \DontPrintSemicolon
	\caption{Fast Minimum-norm (FMN) Attack}
	\label{alg:alg_fmn}
   \Input{$\vct x$, the input sample; $y$, the target (true) class label; $\alpha_{0}$, the initial $\vct \delta$-step size; $K$, the total number of iterations; \textcolor{blue}{$L$, the loss of the attack}; \textcolor{red}{$h$, the step size scheduler}; \textcolor{cyan}{$u$, the update function for the gradient}.}
	\Output{The minimum-norm adversarial example $\vct x^\star$.}
     $\vct x_0 \gets \vct x, \, \epsilon_0 = 0, \, \vct \delta_{0} \gets \vct 0, \, \vct \delta^\star \gets \vct \infty$, $\gamma_{0}=0.05$ \label{line:init}\Comment*[r]{initialization}
    \For {$k = 1, \ldots, K$} 
	    {$\vct g \gets \nabla_{\vct \delta} \textcolor{blue}{L}(\vct x +\vct \delta_{k-1}, y,  \vct \theta)$ \Comment*[r]{\textcolor{blue}{loss} gradient}\label{line:gradient}
	    $\gamma_{k} \gets h_\gamma(\gamma_{0}, k, K)$ \Comment*[r]{$\epsilon$-step size decay} \label{line:gamma_decay}
			$\epsilon_{k} = u_\epsilon(\epsilon_{k-1}, \gamma_k, \|\vct \delta\|_p)$\label{line:eps_step} \Comment*[r]{$\epsilon$-step}
  $\alpha_{k} \gets \textcolor{red}{h}(\alpha_{0}, k, K)$ \Comment*[r]{\textcolor{red}{scheduler} step}\label{line:step_decay}
        $\vct \delta_{k} \gets$ \textcolor{cyan}{$u$}($\vct \delta_{k-1}, \vct g / \|\vct g\|_2 , \alpha_{k}$) \label{line:opt_step} \Comment*[r]{\textcolor{cyan}{optimizer} step}
        $\vct \delta_k \gets \Pi (\vct x_0, \vct \delta_k)$ \label{line:project} \Comment*[r]{projection onto the feasible domain}
		}
	
	\bfseries return $\vct x^\star \gets \vct x_0 + \texttt{best}(\vct \delta_0, ... \vct \delta_K$) \label{line:return} \Comment*[r]{return best solution}
\end{algorithm}

%% file: table.tex
\begin{table}[htbp]
\centering
\caption{Results for HO-FMN, compared with reference values for AA. We highlight in yellow HO-FMN (SGD) and in blue HO-FMN (Adam).}
\label{tab:configurations}
\resizebox{0.99\textwidth}{!}{%
\begin{tabular}{@{}lll|rrrrrrrrr|r@{}}
\toprule
\textbf{Optim.} & \textbf{Sched.} & \textbf{Loss} & \textbf{M0} & \textbf{M1}   & \textbf{M2}   & \textbf{M3}   & \textbf{M4}   & \textbf{M5}   & \textbf{M6}   & \textbf{M7}   & \textbf{M8}   & \textbf{Mean} \\ \midrule
\rowcolor[HTML]{E2EFDA} 
\multicolumn{3}{l|}{AA} & 0.71 & 0.67 & 0.66 & 0.65 & 0.63 & 0.63 & 0.63 & 0.61 & 0.59 & 0.64 \\
\midrule
\rowcolor[HTML]{FFF2CC} 
\multicolumn{1}{l}{SGD} &CALR &LL         & \textbf{0.76}        & 0.74          & 0.70          & 0.66          & 0.66          & 0.68          & 0.66          & 0.58          & 0.62          & 0.67          \\ \midrule
\rowcolor[HTML]{E2EFFF}
\multicolumn{1}{l}{Adam}               & RLROP              & LL            & \textbf{0.76}        & \textbf{0.72}          & \textbf{0.68 }         & \textbf{0.64}          & \textbf{0.58}          & \textbf{0.62}          & \textbf{0.56}          & \textbf{0.56}          & \textbf{0.60}         & \textbf{0.64}          \\ \midrule
\multicolumn{1}{l}{Adam}                & CALR               & LL            & 0.78        & 0.74          & 0.68          & 0.64          & 0.62          & 0.66          & 0.64          & 0.58          & 0.62          & 0.66          \\
\multicolumn{1}{l}{Adam}               & MSLR               & LL            & 0.78        & 0.74          & 0.70          & 0.66          & 0.62          & 0.66          & 0.64          & 0.58          & 0.62          & 0.67          \\
\multicolumn{1}{l}{SGD}                & RLROP              & LL            & 0.76        & 0.74          & 0.70          & 0.64          & 0.64          & 0.68          & 0.64          & 0.58          & 0.62          & 0.67          \\
\multicolumn{1}{l}{Adam}                & CAWR               & LL            & 0.78        & 0.74          & 0.72          & 0.66          & 0.64          & 0.66          & 0.66          & 0.58          & 0.62          & 0.67          \\
\multicolumn{1}{l}{SGD}                & CALR               & LL            & 0.78        & 0.74          & 0.70          & 0.66          & 0.64          & 0.68          & 0.66          & 0.58          & 0.62          & 0.67          \\
\multicolumn{1}{l}{SGD}                & CAWR               & LL            & 0.78        & 0.74          & 0.70          & 0.66          & 0.66          & 0.68          & 0.66          & 0.58          & 0.62          & 0.68          \\
\multicolumn{1}{l}{SGD}                  & MSLR               & LL            & 0.78        & 0.74          & 0.74          & 0.66          & 0.66          & 0.68          & 0.66          & 0.58          & 0.62          & 0.68          \\
\midrule
\multicolumn{1}{l}{Adam}                & MSLR               & CE            & 0.90        & 0.90          & 0.86          & 0.76          & 0.76          & 0.86          & 0.70          & 0.86          & 0.88          & 0.83          \\
\multicolumn{1}{l}{SGD}               & MSLR               & CE            & 0.96        & 0.94          & 0.88          & 0.84          & 0.82          & 0.92          & 0.80          & 0.82          & 0.80          & 0.86          \\
\multicolumn{1}{l}{SGD}                  & CALR               & CE            & 1.00        & 0.96          & 0.90          & 0.88          & 0.86          & 0.94          & 0.86          & 0.88          & 0.84          & 0.90          \\
\multicolumn{1}{l}{SGD}                  & RLROP              & CE            & 1.00        & 0.96          & 0.90          & 0.88          & 0.86          & 0.94          & 0.86          & 0.90          & 0.84          & 0.90          \\
\multicolumn{1}{l}{SGD}                  & CAWR               & CE            & 1.00        & 0.96          & 0.92          & 0.90          & 0.86          & 0.94          & 0.88          & 0.90          & 0.88          & 0.92          \\
\multicolumn{1}{l}{Adam}                & CALR               & CE            & 1.00        & 1.00          & {0.94}     & 0.96          & {0.90}    & {0.96}        & {0.92} & {0.94} & {0.94} & 0.95          \\
\multicolumn{1}{l}{Adam}                & CAWR               & CE            & 1.00        & 1.00          & 0.96          & 0.96          & {0.94} & {0.96}        & {0.92} & {0.94} & {0.94} & 0.96          \\
\multicolumn{1}{l}{Adam}               & RLROP              & CE            & 1.00        & 1.00          & 0.96          & 0.96          & {0.92}         & 0.98          & {0.92} & {0.94}       & {0.94} & 0.96          \\
\bottomrule
\end{tabular}%
}
\end{table}